\pdfoutput=1

\documentclass[11pt]{article}

\usepackage{ACL2023}
\usepackage{graphicx}
\usepackage{caption}
\usepackage{subcaption}
\usepackage{float}
\usepackage{amsmath}
\usepackage{amssymb}
\usepackage{times}
\usepackage{latexsym}
\usepackage{booktabs}
\usepackage{array}
\usepackage{colortbl}

\usepackage[T1]{fontenc}

\usepackage[utf8]{inputenc}

\usepackage{microtype}

\usepackage{inconsolata}

\title{\textit{Residual Prompt Tuning}: Improving Prompt Tuning\\ with Residual Reparameterization}

\author{Anastasia Razdaibiedina$^{\diamondsuit}$\enspace Yuning Mao$^{\spadesuit}$\enspace Madian Khabsa$^{\spadesuit}$ \AND Mike Lewis$^{\spadesuit}$\enspace Rui Hou$^{\spadesuit}$\enspace Jimmy Ba$^{\diamondsuit}$\enspace Amjad Almahairi$^{\spadesuit}$ \\ 
$^{\diamondsuit}$University of Toronto \& Vector Institute \enspace \enspace $^{\spadesuit}$Meta AI \\
\texttt{\{sadalsuud, jba\}@cs.toronto.edu} \\
\texttt{\{yuningm, rayhou, mkhabsa, mikelewis, aalmah\}@meta.com}
}

\begin{document}
\maketitle
\begin{abstract}
Prompt tuning is one of the successful approaches for parameter-efficient tuning of pre-trained language models. Despite being arguably the most parameter-efficient (tuned soft prompts constitute $<0.1\%$ of total parameters), it typically performs worse than other efficient tuning methods and is quite sensitive to hyper-parameters. In this work, we introduce \textsc{Residual Prompt Tuning} -- a simple and efficient method that significantly improves the performance and stability of prompt tuning. We propose to reparameterize soft prompt embeddings using a shallow network with a residual connection. Our experiments show that \textsc{Residual Prompt Tuning} significantly outperforms prompt tuning on SuperGLUE benchmark 
across T5-Large, T5-Base and BERT-Base models. Notably, our method reaches $+7$ points improvement over prompt tuning with T5-Base and allows to reduce the prompt length by $\times 10$ without hurting performance. In addition, we show that our approach is robust to the choice of learning rate and prompt initialization, and is effective in few-shot settings.\footnote{Our code is available at \href{https://github.com/arazd/ResidualPrompts}{https://github.com/\\arazd/ResidualPrompts}. 
}

\end{abstract}

\section{Introduction}

Pre-trained language models have achieved remarkable performance on a variety of natural language understanding tasks \citep{devlin2018bert, liu2019roberta, raffel2020exploring}. Recent studies have shown that scaling up model size consistently leads to performance gains \citep{kaplan2020scaling, raffel2020exploring, zhang2022opt}, and larger scale models are becoming increasingly more common, e.g. GPT-3, 175B parameters \citep{brown2020language}, MT-NLG, 530B parameters \citep{smith2022using}. 
Despite the significant performance improvement achieved with larger-scale models, their applicability is limited due to their size. The standard practice of \textit{fine-tuning} 
becomes prohibitively expensive since it requires storing gradients and optimizer states for all model parameters. Additionally, storing a separate copy of a fine-tuned model for each task is infeasible for billion-parameter models.

\begin{figure}
\centering
\includegraphics[scale=0.37]{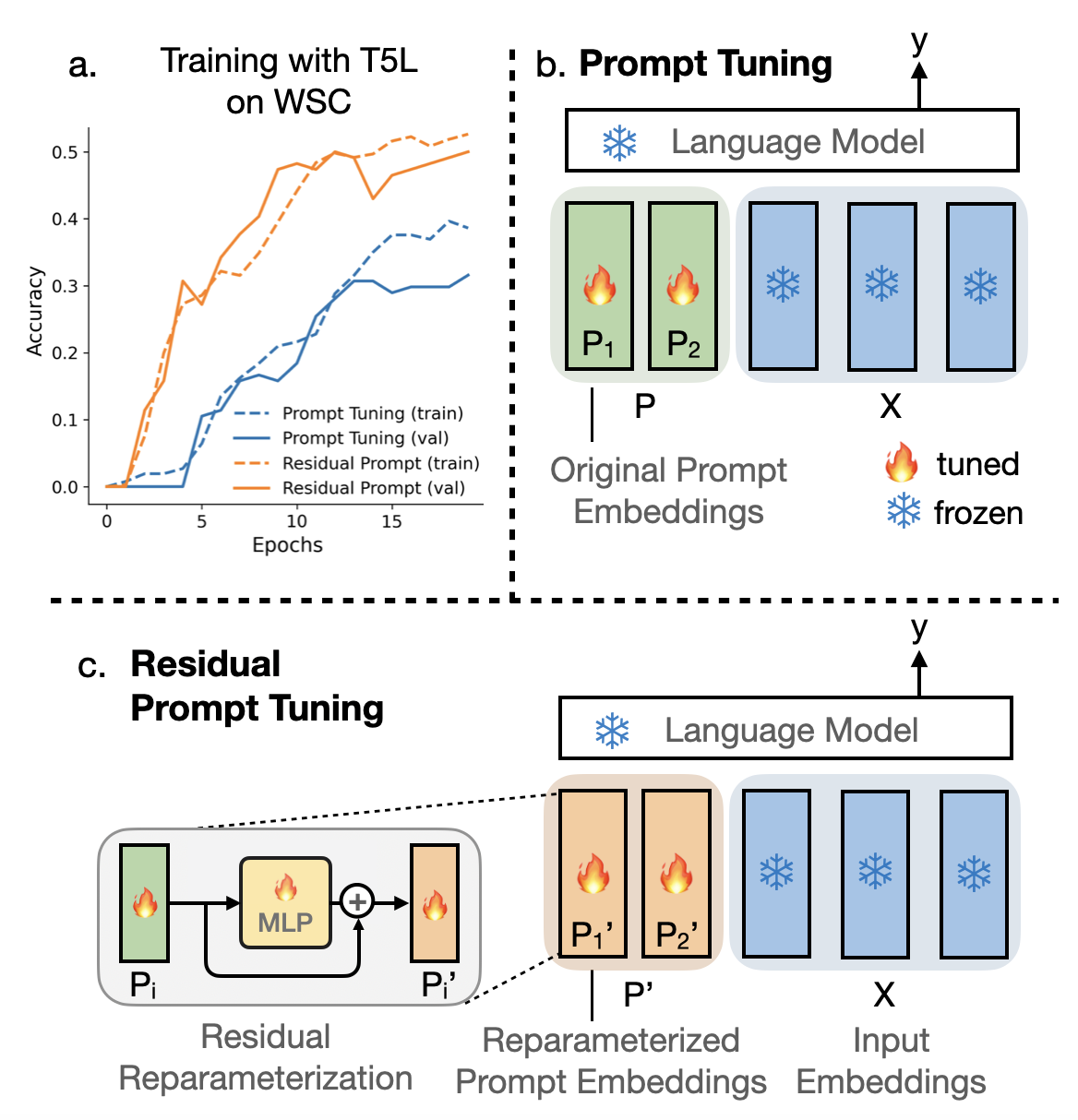}
\caption{Illustration of \textsc{Residual Prompt Tuning} and comparison with prompt tuning by \citet{lester2021power}. \textbf{a.} \textsc{Residual Prompt Tuning} reaches stronger performance than prompt tuning (performance with T5-Large model on WSC task is shown). \textbf{b.} Prompt Tuning tunes prompt embeddings $P$, which are concatenated with input embeddings $X$ and fed into the frozen language model. 
 \textbf{c.} \textsc{Residual Prompt Tuning} passes the original prompt embeddings $P$ through a shallow network (e.g. MLP) with a residual connection and then prepends them to the input. Embeddings $P$ and MLP parameters are jointly tuned.
 }
\label{fig:method}
\end{figure}

To address the challenges associated with full model tuning, a line of research has focused on \textit{prompt design}, where natural language prompts are used to query a frozen model \citep{brown2020language}. In this setup, all tasks are cast as language modeling tasks (e.g. 0/1 classes could be encoded
as ``True``/``False``), and manually selected prompts condition the frozen model to generate the desired output. Despite the fact that prompt design can achieve strong few-shot performance, manually finding optimal prompts remains challenging and time-consuming \citep{zhao2021calibrate}. Additionally, different prompt choices often lead to large variances in the final performance \citep{zhao2021calibrate, vu2021spot}.  

Recently, \citet{lester2021power} proposed \textit{prompt tuning} -- a method of learning \textit{soft prompts} through gradient descent instead of designing the prompts manually. Soft prompts are a series of continuous embeddings prepended to the input, which are updated throughout training, and typically constitute $<0.1\%$ of the total parameters. 
Notably, prompt tuning has been shown to perform close to full model tuning when model size increases, closing the performance gap when the model contains over 11B parameters \citep{lester2021power}. Nevertheless, prompt tuning still underperforms with smaller models, and its performance can vary significantly depending on the choice of hyperparameters, such as prompt initialization and learning rate \citep{vu2021spot}. Furthermore, prompt tuning generally requires long training and a large number of prompt tokens (over 100) to achieve stable performance \citep{lester2021power}. This becomes a major bottleneck when prompts are learned sequentially in a continual learning setup \cite{razdaibiedina2023progressive} or context length is limited.

In this work, we present \textsc{Residual Prompt Tuning}, a method that can significantly improve and stabilize prompt tuning performance through residual reparameterization of prompt embeddings (Figure~\ref{fig:method}). \textsc{Residual Prompt Tuning} passes soft prompt embeddings through a shallow network with a residual connection, and subsequently prepends reparameterized prompt to the input and feeds to the language model. This reparameterization gives the model more flexibility to decide between using a separate embedding for each prompt token versus the representation obtained from the shared reparameterization network. 
After training is completed, the reparameterization network can be discarded and original prompt embeddings can be replaced with their projections.

We conduct extensive experiments on SuperGLUE tasks with T5-Large, T5-Base and BERT-Base models \citep{raffel2020exploring, devlin2018bert} and demonstrate that \textsc{Residual Prompt Tuning} outperforms previous prompt tuning-based methods by a large margin, achieving $+7$ points improvement over prompt tuning on SuperGLUE with T5-Base. We also show that \textsc{Residual Prompt Tuning} reduces performance variance under different learning rates or prompt initializations, and achieves strong performance with fewer training iterations. Finally, we show that \textsc{Residual Prompt Tuning} significantly improves over prompt tuning in few-shot settings. 


\section{Background}

\textbf{Fine-tuning. }
The predominant approach for adapting a pre-trained language model to a downstream task is to fine-tune all its parameters $\Theta$ \citep{devlin2018bert, raffel2020exploring}. Consider a classification task $T$ with input text $x$, and output scalar label $y$, where $p_\Theta$ is a probability
distribution of output classes parameterized by the full model weights $\Theta$. The training objective is simply:
\begin{equation}
\max_{\Theta} \sum_{x,y \in T} \log p_{\Theta}(y|x).
\end{equation}
Despite its effectiveness, fine-tuning updates all model parameters, which can be prohibitively expensive for large language models.
\\
\\
\textbf{Prompt Tuning.} \citet{lester2021power} proposed prompt tuning as a lightweight alternative to fine-tuning. The main idea is to prepend a sequence of virtual token embeddings, or a \textit{soft prompt} $P$, to the input text $x$, and learn only them on the downstream task while keeping other model parameters fixed. 
The model parameters $\Theta$ are now composed of the frozen pre-trained language model parameters, 
and the additional soft prompt parameters $\theta_P$, which are tuned on the downstream task. The training objective becomes:
\begin{equation}
\label{eqn:PT}
\max_{\theta_P} \sum_{x,y \in T} \log p_\Theta(y|[P; x]).
\end{equation}
Prompt tuning offers an attractive parameter-efficient solution to repurpose pre-trained models for many real-world applications. However, training soft prompts often requires extensive hyperparameter tuning and longer training time to achieve the desired performance \citep{lester2021power}.

\section{Method}
\subsection{\textsc{Residual Prompt Tuning}}
\begin{figure}[t]
\centering
\includegraphics[scale=0.3]{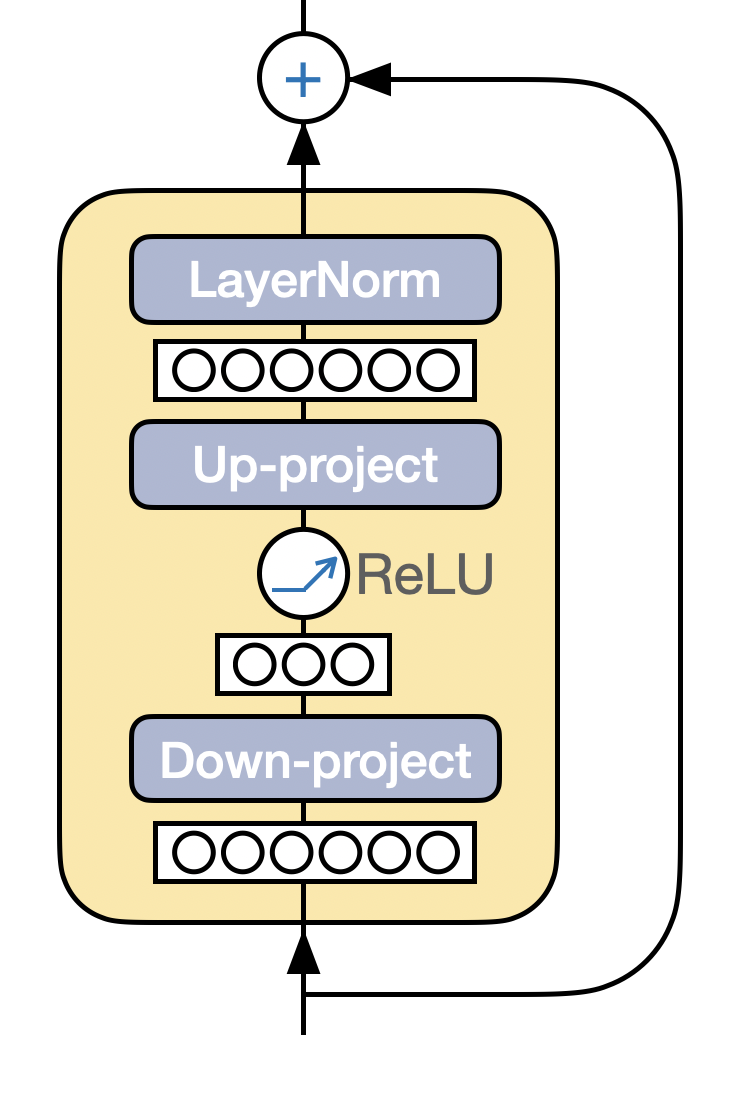}
\vspace{-0.5cm}
\caption{Illustration of reparameterization network $\Phi$ used in \textsc{Residual Prompt Tuning}. Each virtual token $P_i$ is passed through the down-projection layer, followed by the non-linearity, then the up-projection layer, and the normalization layer, and then summed with the unprojected embedding via skip connection.}
\label{fig:mlp}
\end{figure}

We propose to use a more flexible parameterization of soft prompts
using a shallow network with a skip connection (Figure~\ref{fig:method}).
Specifically, we project the sequence of prompt embeddings $P$ consisting of $n$ virtual tokens $[P_1, ..., P_n]$ into a reparameterized sequence $P'$ as follows:
\begin{equation}
    P' = [P'_1, ..., P'_n] = [\Phi(P_1), ..., \Phi(P_n)],
\end{equation}
where $\Phi(\cdot)$ is a reparameterization function composed of a shallow network $\phi(\cdot)$ with a residual connection. $\Phi(\cdot)$ is applied independently to each prompt token:
\begin{equation}
    \Phi(P_i) = \phi(P_i) + P_i,\; i \in \{1...n\}
\end{equation}
Our $\phi(\cdot)$ network is a multi-layer perceptron (MLP) that follows a "bottleneck" design, as in commonly used ResNet building blocks \citep{he2016deep} and adapter modules \citep{houlsby2019parameter}.  
It consists of down-projection $\mathbf{W}_{\text{down}} \in \mathbb{R}^{d \times m}$ and up-projection $\mathbf{W}_{\text{up}} \in \mathbb{R}^{m \times d}$ layers (as shown in Figure~\ref{fig:mlp}), a combination of which has been thoroughly explored in literature \citep{he2016deep, houlsby2019parameter}. Here, $d$ is the dimensionality of model embeddings and $m$ is the bottleneck size of the MLP (hyperparameter of our approach).
We train only the prompt embeddings $\theta_P$ and the repremeterization parameters $\theta_\phi$ on the downstream task, while keeping all other parameters frozen. The training objective is to maximize the log-likelihood of correct output $y$ given the input text $x$ concatenated with the reparameterized soft prompt $P'$:
\begin{equation}
\label{eqn:resPT}
\max_{\theta_P, \theta_{\phi}} \sum_{x,y \in T} \log p_{\Theta}(y|[P'; x]).
\end{equation}

\subsection{Design choices}
We discuss here several important design choices for the reparameterization network $\Phi$.

\textbf{Residual connection.} We find that residual connection plays a key role in boosting performance and speeding up the convergence in \textsc{Residual Prompt Tuning} (Section~\ref{section:main_results}, Appendix~\ref{appendix:pt_res_mlp}). Similar to ResNets \citep{he2016deep}, we hypothesize that  residual learning gives the model more flexibility to decide between using a separate embedding for each prompt token versus the representation obtained from the shared network. 
We discuss further benefits of residual connection in Appendix~\ref{appendix:pt_res_mlp}.

\textbf{Depth and width of MLP.} We use two-layer MLP, whose up- and down-projection matrices $\mathbf{W}_{\text{up}}$ and $\mathbf{W}_{\text{down}}$ constitute the additional trainable parameters. 
Increasing the dimensionality $m$ of the hidden layer results in higher performance (see Section~\ref{sec:ablations}), suggesting that the overparameterization~\citep{allen2019convergence} of prompt tokens is important for the performance improvement. More details on parameter-efficiency are in Appendix~\ref{appendix:efficiecy}.

\textbf{Non-linearity and normalization.} We select LayerNorm \citep{ba2016layer} as our normalization layer and ReLU as our non-linearity. We find that LayerNorm helps to stabilize the performance, while the effect of the specific choice of the non-linear layer is of lesser importance.

\textbf{Parameter sharing.} In our setup, we apply a shared reparameterization network $\Phi$ to each virtual token embedding. Another design choice is to apply a separate network to each prompt embedding. We compare both variants in Section~\ref{sec:ablations}. Overall, a shared MLP is significantly more parameter-efficient and offers the benefit of knowledge sharing in limited data settings.

\subsection{Training and Inference}
During training, we jointly optimize prompt embeddings $P$ and parameters of the reparameterization network $\Phi(\cdot)$, while keeping the backbone model frozen. The reparameterized prompt is inserted before the input text embeddings and fed into the language model (
see details in Section~\ref{sec:architectures}). Importantly, we use task-specific prompts, meaning that reparameterized prompt embeddings are not dependent on the input. 

After training is complete, we project prompt embeddings through the learned reparameterization network $\Phi(\cdot)$, and replace the original prompt embeddings with their corresponding projections $P' = \Phi(P)$. \textbf{During inference, we discard the reparameterization network} and solely use the projected prompt embeddings $P'$. Specifically, we insert $P'$ in front of the input text embeddings, and feed them together to the frozen pre-trained model.

\section{Experiments}

\subsection{Datasets}
Following previous works on prompt tuning \citep{lester2021power, vu2021spot}, we use NLU tasks from the SuperGLUE benchmark to assess the performance of the language model \citep{wang2019superglue}. Specifically, we use the following 8 datasets: BoolQ \citep{clark2019boolq}, CB \citep{de2019commitmentbank}, COPA \citep{roemmele2011choice}, MultiRC \citep{khashabi2018looking}, ReCoRD \citep{zhang2018record}, RTE \citep{giampiccolo2007third}, WiC \citep{pilehvar2018wic} and WSC \citep{levesque2012winograd}. More details on are discussed in Appendix~\ref{appendix:implementation}, ~\ref{appendix:datasets}.

\subsection{Architectures}
\label{sec:architectures}
\textsc{Residual Prompt Tuning} is a model-agnostic approach that can be used with any transformer architecture -- similarly to the original prompt tuning \citep{lester2021power}. In our experiments, we explore the performance of our method with encoder-decoder T5 model\footnote{While \citet{lester2021power} reports better performance with T5 v1.1 compared to T5, several works find version v1.1 less stable for prompt tuning compared to the original T5 and report worse performance \citep{karimi2021compacter, asai2022attentional}. Thus, in this work we use the
original T5 model.}
\citep{raffel2020exploring} and encoder-only BERT model \citep{devlin2018bert}. Specifically, we focus on BERT-Base (110M parameters), T5-Base (220M parameters) and T5-Large (770M parameters) model variants.
\\
\\
\textbf{BERT.} For BERT experiments, we insert the trainable prompt in front of the input sequence, but before the $\mathtt{[CLS]}$ token, resulting in the following input $\hat{x}$ to the language model: 
$\hat{x} = \mathtt{concat[} 
\mathbf{E}( \mathtt{[CLS]}), 
 P', 
\mathbf{E}(S  \mathtt{[EOS]}) 
\mathtt{]},$
where $P'$ is the embeddings matrix of the reparameterized soft prompt, $S$ is the input sentence, $\mathtt{[CLS]}$ and $\mathtt{[EOS]}$ denote special tokens (for sentence classification and marking end-of-sentence), and $\mathbf{E}$ denotes tokenization and extraction of embeddings.

To predict the class of input text $\hat{x}$, we follow the original \citep{devlin2018bert} setup and use encoder representation of the $\mathtt{[CLS]}$  token, $h_{\mathtt{[CLS]}}$, 
and add a linear transformation parameterized by $\mathbf{w}$ and a softmax layer to predict the class of $\hat{x}$:
$$ p(y=c | h) = \frac{ e^{\mathbf{w_c}h_{\mathtt{[CLS]}}} }{ \sum_{k \in \mathcal{C} }{ e^{\mathbf{w_k}h_{\mathtt{[CLS]}}} }} $$
\\ 
After that, we apply cross-entropy loss to perform gradient updates on the prompt embeddings, linear head, and reparameterization network.
\\
\\
\textbf{T5.} For T5 experiments we cast all tasks as language modeling tasks, following \citet{raffel2020exploring, lester2021power}. In this setup, we model the classification task as conditional generation, where output 
is a sequence of tokens that represent a class label. We prepend reparameterized prompt embeddings $P'$ in front of the input text embeddings, hence total input $\hat{x}=\mathtt{concat}[P', \mathbf{E}(S)]$ is passed into the pre-trained language model. 
T5 model applies a multi-headed self-attention over the input tokens followed
by position-wise feed-forward layers to output a distribution over target tokens. We train prompt embeddings and parameters of the reparameterization network with cross-entropy loss. More details on input preprocessing and prompt initialization are in Appendix~\ref{appendix:preprocessing}, ~\ref{appendix:prompt_init}.

\subsection{Baselines}
We compare \textsc{Residual Prompt Tuning} (Res PT) with approaches from two different categories: methods for \textit{prompt reparameterization} and \textit{parameter-efficient tuning} (PEFT) methods. 

In our first set of experiments, we study how much residual reparameterization can improve prompt tuning performance and evaluate it against other reparameterization techniques. In sum, we compare our approach with the original prompt tuning (PT; no reparameterization 
\citealt{lester2021power}), prompt tuning with MLP reparameterization (PT w/ MLP; \citealt{li2021prefix}), prompt tuning with LSTM reparameterization (PT w/ LSTM; \citealt{liu2021gpt}) and fine-tuning.

\begin{table*}[!htbp]
\vspace{-0.3in}
\newcolumntype{g}{>{\columncolor{black!15!white}}c}
\setlength\extrarowheight{-2.1pt}
  \centering
  \scalebox{0.85}{
  \begin{tabular}{lccccccccg}
  \toprule
   Task $\rightarrow$  & BoolQ & CB & COPA & MultiRC & ReCoRD & RTE & WiC & WSC & Avg. \\
   Method $\downarrow$  & Acc.  & F1/Acc.  & Acc.  & F1/EM  & F1/EM & Acc. & Acc.  & Acc. & - \\
\toprule \multicolumn{10}{c}{\textbf{T5-Large}} \\
Prompt Tuning &   83.4 &  86.4 &  54.0 &     67.9 & \textbf{73.3} & \textbf{86.4} &  67.5 &  31.0 &  68.7 \\
PT w/ MLP &   83.4 &  82.1 &  37.0 &     67.9 &    68.8 &           77.4 &  66.2 &   7.0 &  61.2 \\
PT w/ LSTM &  53.8 &  78.9 &   0.0 &     66.4 &    82.1 & 49.5 &  15.2 &   0.0 &  43.2 \\
Residual PT &  \textbf{83.5} &  \textbf{86.9} &  \textbf{56.3} & \textbf{68.6} &    68.1 &           86.2 &  \textbf{70.8} &  \textbf{50.4} &  \textbf{71.4} \\
%

\midrule
Fine-tuning\textsuperscript{\dag} &  85.4 & 93.2 & 83.4 & 67 & 86.3 & 87.8 & 69.3 & 86.3 & 82.3 \\
\toprule \multicolumn{10}{c}{\textbf{T5-Base}} \\

Prompt Tuning &   \textbf{78.0} &  77.4 &  \textbf{58.3} &     59.2 &    59.5 &           63.7 &  66.2 &  37.7 &  62.5 \\
PT w/ MLP &   77.5 &  74.8 &  57.7 &  \textbf{59.5} & \textbf{60.8} & 56.0 &  65.2 &  39.5 &  61.4 \\
PT w/ LSTM &  51.1 &   5.0 &   3.5 &           12.5 &  32.3 &   43.3 &     54.9 &    43.1 &  30.7 \\
Residual PT &   77.9 &  \textbf{79.2} &  \textbf{58.3} & 59.3 &  60.2 & \textbf{70.4} &  \textbf{66.8} &  \textbf{49.1} &  \textbf{65.2} \\
    
\midrule
    Fine-tuning\textsuperscript{\dag} & 81.4 & 86.2 & 94.0 & 71.2 & 61.4 & 74.6 & 68.3 & 80.8 & 76.2 \\
    \toprule \multicolumn{10}{c}{\textbf{BERT-Base}} \\
    Prompt Tuning &   62.2 &  60.7 &  51.6 &     57.5 &    60.0 &  53.1 &  54.3 &  61.9 &  57.7 \\
PT w/ MLP & 62.0 &  61.3 &  53.2 & 58.3 &  \textbf{62.8} &  48.0 &  54.6 &  \textbf{64.1} &  58.0 \\
PT w/ LSTM &   62.2 &  65.2 &  52.0 & 53.1 & 62.7 &  44.6 &  \textbf{59.9} &  63.5 &  57.9  \\
Residual PT &  \textbf{62.7} &  \textbf{67.9} &  \textbf{63.5} & \textbf{59.0} &  61.1 &  \textbf{54.9} &  57.1 &  63.5 &  \textbf{61.2} \\
\midrule
Fine-tuning & 73.2 &  89.9 &  65.7 & 66.9 &    62.8 &  65.1 &  67.8 &  63.8 &  69.4 \\
  \bottomrule
  \end{tabular}
  }
  \caption{Results on SuperGLUE development set with \textit{\textbf{10-token prompt}}. All scores are averaged over 3 runs. \textsuperscript{\dag}denotes results reported by \citet{raffel2020exploring}. For tasks with two metrics, the average score is reported. \vspace{-0.05in}}
  \label{tab:1}
\end{table*} 

In our second set of experiments, we assess the benefits of \textsc{Residual Prompt Tuning} method versus existing PEFT approaches. 
In addition to prompt tuning, we include a set of PEFT baselines: Adapter \citep{houlsby2019parameter}, AdapterDrop \citep{ruckle2020adapterdrop}, SPoT \citep{vu2021spot}, ATTEMPT \citep{asai2022attentional}. Adapter and AdapterDrop approaches are based on adapters by \citet{houlsby2019parameter}, whereas SPoT and ATTEMPT are tranfer learning-based methods for prompt tuning, which find optimal prompt initializations by pre-training prompts on informative source tasks.

\subsection{Experimental setup}
For all experiments with prompt tuning-based methods, we follow standard protocol by \citet{lester2021power} and report results on the validation set. Unless otherwise specified, we use standard metrics associated with each task to report final performance (see Table~\ref{tab:datasets_explained}). For experiments where we compare \textsc{Residual Prompt Tuning} with PEFT methods (Section~\ref{section:subsection_peft}), we follow PEFT training protocol \citep{asai2022attentional, karimi2021compacter}. More experimental details are in Appendix~\ref{appendix:training_details}.


\vspace{0.3cm}
\begin{table}[h]
\scalebox{0.85}{
\addtolength{\tabcolsep}{-3.5pt}    
\begin{tabular}{ lccc|ccc }
\toprule 
Prompt len. $\rightarrow$ & \multicolumn{3}{c}{\textbf{10 tokens}} & \multicolumn{3}{c}{\textbf{100 tokens}}\\ 
 Method $\downarrow$ 
 & T5L & T5B & BERT & T5L & T5B & BERT \\
\midrule 
Prompt Tuning & 68.7 & 62.5 & 57.7 & \textbf{74.5}\textsuperscript{\ddag} & 63.1\textsuperscript{\ddag} & 59.2 \\
PT w/ MLP & 61.2 & 61.4 & 58.0 & 67.8 & 62.4 & 60.8 \\
PT w/ LSTM & 43.2 & 30.7 & 57.9 & 60.1 & 55.2 & 58.8 \\
Residual PT & \textbf{71.4} & \textbf{65.2} & \textbf{61.2} & \textbf{74.5} & \textbf{70.5} & \textbf{61.6} \\
\midrule
Fine-tuning & 82.3\textsuperscript{\dag} & 76.2\textsuperscript{\dag} & 69.4 & 82.3\textsuperscript{\dag} & 76.2\textsuperscript{\dag} & 69.4\\
 \bottomrule 
\end{tabular}
}
\caption{\textsc{Residual Prompt Tuning} outperforms other prompt tuning variations across three architectures (T5-Large, T5-Base and BERT-Base) and different prompt sizes (10 and 100 tokens). Average performance across all SuperGLUE tasks is reported. All our results are averaged over 3 runs. \textsuperscript{\dag}denotes results reported by \citet{raffel2020exploring}; \textsuperscript{\ddag}denotes results reported by \citet{lester2021power, vu2021spot}. \vspace{-0.1in}} \label{tab:superglue_avg}
\end{table}

\section{Results}
We describe our main results showing the effectiveness of \textsc{Residual Prompt Tuning} compared to other prompt tuning-based methods and parameter-efficient methods
in Section~\ref{section:main_results}. We study the robustness of our method to the choice of hyperparameters in Sections~\ref{section:lr} and~\ref{section:init}. Then, we explore the performance of \textsc{Residual Prompt Tuning} in more extreme settings, including smaller prompt sizes (Section~\ref{section:prompt_len}) and few-shot data regime (Section~\ref{section:fewshot}).

\subsection{Main results}
\label{section:main_results}
\subsubsection{Comparison with prompt tuning}
We compare \textsc{Residual Prompt Tuning} with the original prompt tuning, as well as two different reparameterization methods (via MLP and LSTM). Table~\ref{tab:1} shows results for each task with 10-token prompts, and results for 100-token prompts are presented in Appendix~\ref{appendix:table100token}. We perform experiments with T5-Large, T5-Base, and BERT-Base model architectures, and with two different prompt sizes: 10 and 100 tokens. Additionally, we include full model tuning results as an upper-bound performance.

Table~\ref{tab:superglue_avg} summarizes the average performance on SuperGLUE with 10-token and 100-token prompts across three model variants. \textsc{Residual Prompt Tuning} outperforms 
other methods, gaining $+3$ points improvement with 10-token prompts on both T5B and T5L models, and over $+7$ points improvement with 100-token prompts on T5B.

Table~\ref{tab:1} dissects the performance with 10-token prompts, showing per-task results for all SuperGLUE tasks across three model variants. \textsc{Residual Prompt Tuning} leads to consistent improvement over prompt tuning across different tasks. LSTM-based reparameterization shows worse performance compared to our approach. Prompt tuning with MLP reparameterization 
experiences significant fluctuations depending on the task -- with stronger performance on ReCoRD ($+0.6$ points), but substantially lower score on WiC ($-9.6$ points) compared to our approach. Overall, \textsc{Residual Prompt Tuning} shows strong improvement over prompt tuning and other reparameterization methods across all model architectures.

\begin{figure}[h]
\vspace{-0.1in}
\centering
\includegraphics[scale=0.4]{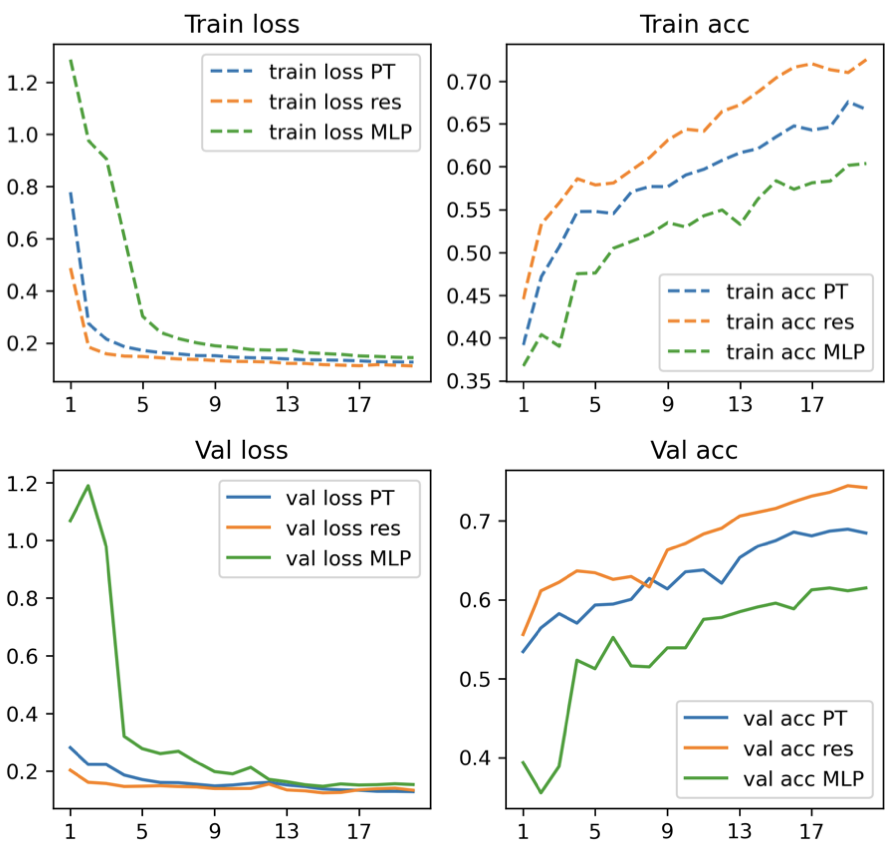}
\vspace{-0.1in}
\caption{\textsc{Residual Prompt Tuning} (orange) speeds up the optimization process of prompt embeddings compared to \textit{prompt tuning} (blue) and \textit{prompt tuning with MLP} (green). 
The x-axis shows the number of training epochs, the y-axis shows loss or accuracy on the train/development sets of RTE. Each point represents an average of 3 runs of T5B model with 10-token prompt. \vspace{-0.1in}}
\label{fig:convergence_rte}
\end{figure}
As shown in Figure~\ref{fig:convergence_rte}, \textsc{Residual Prompt Tuning} leads to faster convergence compared to other methods. The residual connection in the reparameterization
network plays a key role in boosting performance
– MLP-based reparameterization without skip connection leads to slower converge than prompt tuning. We hypothesize that skip connection allows to bypass learning the identity function, and learns projections "on top" of
the original embeddings (similar observations by \citet{he2016deep}. 
We discuss convergence in more detail in Appendix~\ref{appendix:pt_res_mlp}.

\subsubsection{Other parameter-efficient methods}
\label{section:subsection_peft}
We compare the performance of different PEFT methods 
on SuperGLUE benchmark. Here, for all the experiments, we follow \citet{asai2022attentional} setup and train T5-Base model with a 100-token prompt on a selection of 5 SuperGLUE tasks (details in Appendix~\ref{appendix:training_details}). Our results are shown in Table~\ref{tab:peft}.

Notably, \textsc{Residual Prompt Tuning} achieves significant performance gains over prompt tuning, achieving over $+10$ points improvement in average score. A major benefit of our method is that it does not require transfer learning on source tasks to achieve strong results, contrary to two other prompt tuning-based methods: SPoT and ATTEMPT. \textsc{Residual Prompt Tuning} substantially outperforms SPoT ($+6.1$ points), and reaches close performance to ATTEMPT ($1.5$ points difference) without being pre-trained on any source tasks. 

\begin{table}[!h]
\begin{subtable}{1\linewidth}
\newcolumntype{g}{>{\columncolor{black!15!white}}c}
\scalebox{0.85}{
\addtolength{\tabcolsep}{-2.7pt}    
\begin{tabular}{ lcccccg }
\toprule 
 Task $\rightarrow$ & CB & Bool & Multi & WiC & WSC & Avg. \\
 Method $\downarrow$ & F1 & Acc. & F1 & Acc. & Acc. & Avg. \\
 \midrule 
 Fine-tune\textsuperscript{$\ast$} & 85.7 & 81.1 & 72.8 & 70.2 & 59.6 & 73.9  \\ 
 Adapter\textsuperscript{$\ast$} & 85.7 & 82.5 & 75.9 & 67.1 & 67.3 & 75.7     \\
 AdaptDrop\textsuperscript{$\ast$} & 85.7 & 82.3 & 72.9 & 68.3 & 67.3 & 75.3 \\
 ATTEMPT\textsuperscript{$\ast$} & 78.6 & 78.8 & 74.4 & 66.8 & 78.6 & 70.5 \\
 SPoT\textsuperscript{$\ast$} & 46.4 & 77.2 & 74.0 & 67.0 & 50.0 & 62.9  \\
 PT\textsuperscript{$\ast$} & 67.9 & 61.7 & 58.7 & 48.9 & 51.9 & 57.8 \\
 Res-PT &  86.0 &   79.0 &     58.9 &  68.4 &  52.6 &  69.0  \\
 \bottomrule 
\end{tabular}
}
\caption{Performance comparison across five SuperGLUE tasks averaged over 3 runs. \textsuperscript{$\ast$} denotes results reported by \citet{asai2022attentional}. \vspace{-0.1in}} \label{tab:peft}
\vspace{0.5cm}
\newcolumntype{g}{>{\columncolor{black!15!white}}c}
\scalebox{0.85}{
\addtolength{\tabcolsep}{-2.7pt}    
\begin{tabular}{ lcccccg }
\toprule 
 Method & Extra params & Train. params & Pre-train.  \\
 \midrule 
 Fine-tune & - & 220M & \textbf{No} \\ 
 Adapter & 1.9M &  1.9M & \textbf{No}  \\
 AdaptDrop & 1.1M & 1.1M & \textbf{No} \\
 ATTEMPT & 223K & 223K & Yes \\
 SPoT & \textbf{77K} & \textbf{77K} & Yes \\
 PT & \textbf{77K} & \textbf{77K} & \textbf{No}\\
 Res-PT & \textbf{77K} & 462K & \textbf{No} \\
 \bottomrule 
\end{tabular}
}
\caption{Comparison of parameter-efficiency. \textit{Train. params} denotes total number of trainable parameters, \textit{Add. params} denotes number of additional parameters that would be injected into the language model during inference, \textit{Pre-train} denotes if the method requires pre-training on source tasks.}
\label{tab:peft_appendix}
\end{subtable}%
\caption{Comparison of \textsc{Residual Prompt Tuning} (Res-PT) with other parameter-efficient tuning methods.}
\end{table}

\begin{figure*}[h]
\vspace{-0.35in}
\centering
\includegraphics[scale=0.47]{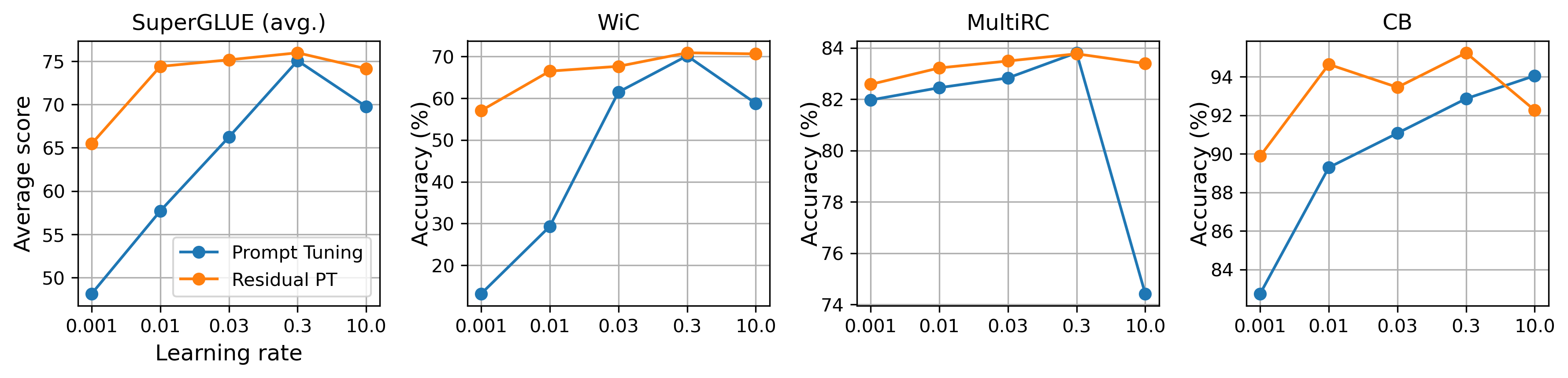}
\vspace{-0.1in}
\caption{Robustness of \textsc{Residual Prompt Tuning} to the choice of learning rate. Experiments are performed with T5L model and 100-token prompt. The x-axis shows different learning rates, the y-axis represents the corresponding performance on the development set. First plot shows average performance on SuperGLUE with respect to the learning rate; other figures show performance on three SuperGLUE tasks (WiC, MultiRC and CB). \vspace{-0.1in}}
\label{fig:learning_rates}
\end{figure*}

Table~\ref{tab:peft_appendix} shows details on the parameter and training efficiency of the explored PEFT methods. Since parameters of the reparameterization network are discarded after training, our method requires the same number of additional parameters at inference as the original prompt tuning. Notably, \textsc{Residual Prompt Tuning} requires 25 times fewer extra parameters during inference compared to adapter-based methods. Finally, our methods does not require pre-training on any source tasks. For more details on parameter-efficiency of \textsc{Residual Prompt Tuning}, see Appendix~\ref{appendix:efficiecy}.



\subsection{Robustness to the choice of learning rate}
\label{section:lr}
We study the performance of \textsc{Residual Prompt Tuning} across a wide range of learning rates (Figure~\ref{fig:learning_rates}). Previous works report that prompt tuning is very sensitive to the learning rate and requires extensive hyperparameter search to reach optimal performance \citep{lester2021power, vu2021spot}. 

We evaluate the performance of our proposed approach and prompt tuning \citep{lester2021power} with learning rates from $\{0.001, 0.01, 0.03, 0.3, 10\}$ on SuperGLUE benchmark. For fair comparison, we use the most stable model variant: T5-Large model with 100-token prompt. Our results are shown in Figure~\ref{fig:learning_rates}. Notably, residual reparameterization allows stabilizing prompt tuning performance across a wide range of learning rates. Original prompt tuning often experiences fluctuations in its performance, with some tasks favoring lower learning rates (e.g. MultiRC), other tasks performing better with higher learning rates (e.g. CB), and yet other tasks achieving peak performance at a specific learning rate (e.g. WiC). In contrast to prompt tuning, \textsc{Residual Prompt Tuning} is robust to the choice of learning rate -- it achieves strong performance with minimal fluctuations (less than $2$ points on average SuperGLUE score) with learning rates between 0.01 and 10 (over 100-fold variation).

\subsection{Robustness to the prompt initialization}
\label{section:init}
\citet{lester2021power} finds that initialization of prompt parameters plays a major role in the final performance. Specifically, initializing prompt embeddings from sampled vocabulary embeddings can boost average SuperGLUE performance by up to $+10$ points compared to random uniform initialization \citep{lester2021power}. Here, we asked if \textsc{Residual Prompt Tuning} performance would depend on the choice of initialization.

Table~\ref{table:init} shows our results (initialization details are in Appendix~\ref{appendix:prompt_init}; we use T5B model with 10-token prompt). We can see that \textsc{Residual Prompt Tuning} is robust to the prompt initialization method, reaching comparable results with both initialization choices: $0.8$ points average performance difference between random uniform initialization and sampled vocabulary initialization. Of note, the initialization effect is more pronounced for smaller-scale dataset CB (250 samples) -- random initialization attributes to $-0.3$ performance drop for \textsc{Residual Prompt Tuning} versus $-4.5$ score difference for the original prompt tuning.

\begin{center}
\begin{table}[h]
\scalebox{0.79}{
\newcolumntype{g}{>{\columncolor{black!15!white}}c}
\addtolength{\tabcolsep}{-2.7pt} 
\begin{tabular}{ lcccccg } 
 \toprule
 Task $\rightarrow$ & & CB & WiC & Multi & RTE & Avg. \\
 Method $\downarrow$ & Init. $\downarrow$ & F1/Acc & Acc & F1/Acc & Acc & - \\
 \midrule

Prompt tune & Rand. &   72.9 &  65.0 &     59.1 &  63.7 &  65.2 \\
Prompt tune & Vocab. &  77.4 &  66.2 &     59.2 &  63.7 &  66.6 \\
delta & - & 4.5 &   1.2 &      0.1 &   0.0 &   1.5 \\
\midrule
Res-PT & Rand. &   78.9 &  66.8 &     59.4 &  67.3 &  68.1 \\
Res-PT & Vocab. &  79.2 &  66.8 &     59.3 &  70.4 &  68.9 \\
delta & - &   0.3 &   0.0 &     -0.1 &   3.1 &   0.8 \\
 \bottomrule
\end{tabular}
}
\caption{Robustness of \textsc{Residual Prompt Tuning} the prompt initialization method. We show results for two prompt  initialization methods: sampled uniformly from the range $[-0.5, 0.5]$ (Rand.), and initializing from the sampled vocabulary (Vocab.). We use T5B model and 10-token prompt. Delta denotes the performance difference between two initialization choices. \vspace{-0.2in}}
\label{table:init}
\end{table}
\end{center}

\begin{figure*}[!htb]
\vspace{-0.25in}
\centering
\includegraphics[scale=0.42]{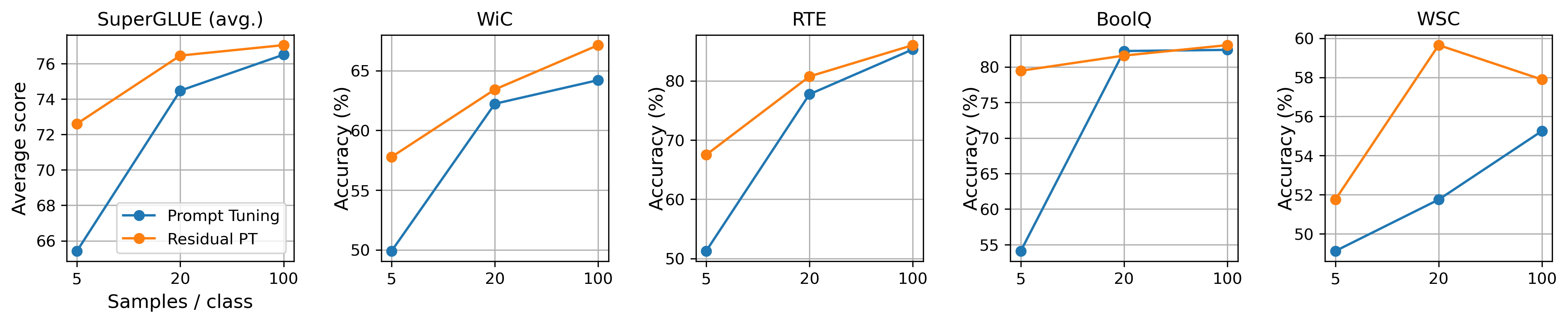}
\vspace{-0.1in}
\caption{Comparison of \textsc{Residual Prompt Tuning} to the prompt tuning in few-shot setting (5, 20, and 100 samples/class) using T5L model and 100-token prompt. The x-axis shows the number of samples per class, the y-axis represents the corresponding performance on the development set. \textit{Left corner} shows average performance on SuperGLUE with respect to the $k$-shot training setup; other figures show performance on specific SuperGLUE tasks (WiC, RTE, BoolQ, and WSC).\vspace{-0.1in}}
\label{fig:fewshot}
\end{figure*}

\subsection{Prompt tuning in few-shot setting}
\label{section:fewshot}
We perform further experiments in few-shot settings (Figure~\ref{fig:fewshot}). Specifically, we sample 5, 20, and 100 samples per class. To avoid variance due to selected samples, we fix the same training subset across all runs for each task; we use T5-Large model and 100-token prompt (as it reaches strongest performance for prompt tuning baseline). 
\textsc{Residual Prompt Tuning} is very effective in few-shot setup, boosting prompt tuning performance by $+7$ and $+2$ points on SuperGLUE benchmark with 5 and 20 samples per class.

\subsection{Performance and prompt length}
\label{section:prompt_len}
We evaluate the \textsc{Residual Prompt Tuning} performance with smaller prompt sizes, and compare it to the original prompt tuning by \citet{lester2021power}. Specifically, we explore the performance with prompts of lengths 2, 10, and 100 tokens with T5-Large model. Our results are shown in Table~\ref{table:prompt_size}. In sum, \textsc{Residual Prompt Tuning} improves performance across all prompt lengths over prompt tuning, achieving average improvement of $+2.6$, $+1.1$ and $+0.8$ points with 2, 10, and 100-token prompts correspondingly.

\begin{table}[h]
\scalebox{0.8}{
\newcolumntype{g}{>{\columncolor{black!15!white}}c}
\addtolength{\tabcolsep}{-1.8pt} 
\begin{tabular}{llccccg}
\toprule
Prompt &  & CB &   WiC &  Multi &  RTE &  Avg.\\
Len. $\downarrow$ & Method $\downarrow$ & Acc & Acc & F1/Acc & Acc & -   \\
\midrule
2 & PT & 91.7 &  67.4 &     84.8 &           81.0 &  81.2 \\
2 &  Res-PT & \textbf{94.0} &  \textbf{70.7} & \textbf{84.9} & \textbf{85.6} & \textbf{83.8} \\
\midrule
10 &  PT & 92.9 &  67.7 &     85.0 & \textbf{86.4} &  83.0 \\
10 &  Res-PT & \textbf{94.0} &  \textbf{71.0} & \textbf{85.1} & 86.2 &  \textbf{84.1} \\
\midrule
100 &  PT & 92.9 &  70.2 & \textbf{83.8} & \textbf{87.5} &  83.6 \\
100 &  Res-PT & \textbf{95.2} &  \textbf{71.3} & \textbf{83.8} & 87.0 & \textbf{84.4} \\
\bottomrule
\end{tabular}
}
\caption{Comparison of \textsc{Residual Prompt Tuning} and prompt tuning by \citet{lester2021power} across different prompt lengths (2, 10, 100 tokens) with T5L model.}
\label{table:prompt_size} 
\end{table}

\subsection{Ablation studies}
\label{sec:ablations}
\textbf{Parameter sharing.} We ablate the effect of a shared reparameterization network by assessing the performance when each prompt is reparameterized through a separate MLP with a skip connection (Table~\ref{tab:ablation_shared}). 
\begin{table}[h]
\newcolumntype{g}{>{\columncolor{black!15!white}}c}
\addtolength{\tabcolsep}{-2pt} 
    \centering
    \scalebox{0.85}{
    \begin{tabular}{lccccg}
    \toprule
    {} & CB & COPA & WiC &  RTE  &  Avg. \\
     & Acc. & Acc. & Acc. & Acc. & -\\
    \midrule
    shared MLP &  \textbf{83.1} &  58.7 &  66.7 &           71.6 &  70.0 \\
separate MLPs &  81.1 &  \textbf{60.3} &  \textbf{67.8} &           \textbf{74.5} &  \textbf{70.9} \\
    \bottomrule
    \end{tabular}
    }
    \caption{Performance of \textsc{Residual Prompt Tuning} with \textit{shared} and \textit{separate} embedding reparameterization networks on four SuperGLUE tasks with T5B.}
    \label{tab:ablation_shared}
\end{table}
\\
We select four SuperGLUE tasks of different sizes: small-scale CB and COPA (250 and 400 training examples), and larger-scale WiC and RTE (6,000 and 2,500 training examples). Interestingly, shared reparameterization network is beneficial in the low data regime, outperforming separate networks by $+2$ points on CB dataset. However, on larger datasets separate networks achieve slightly better performance at the expense of more trained parameters. We show more detailed results in Appendix~\ref{appendix:ablations_shared}.

\begin{figure}[h]
\centering
\includegraphics[scale=0.495]{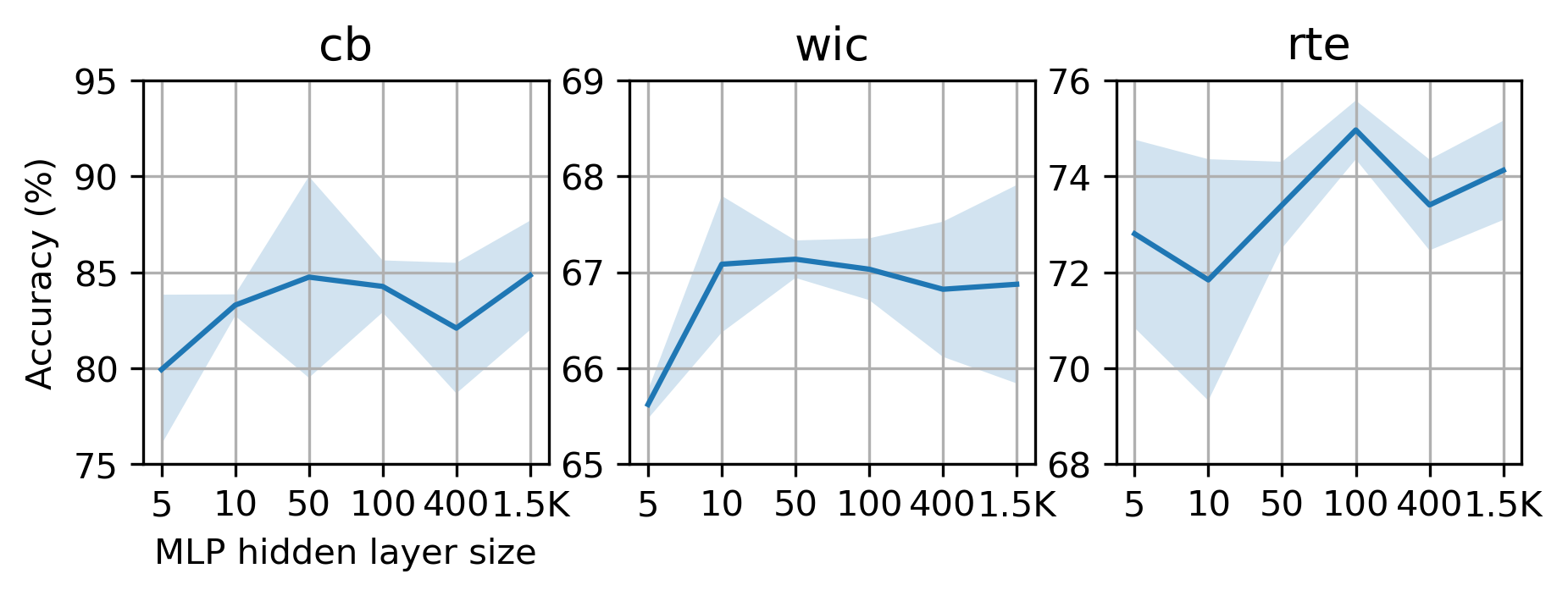}
\vspace{-0.3in}
\caption{The effect of MLP hidden layer size on the performance of \textsc{Residual Prompt Tuning} with T5B. The x-axis shows the hidden layer size, the y-axis shows the maximal validation performance. Each result is averaged over 3 runs, shadow depicts the standard deviation.}
\label{fig:ablation_MLP_size}
\end{figure}

\textbf{Overparameterization.} To study the effect of overparameterization on the final performance, we ablate MLP width by varying the dimension of MLP hidden layer in the following range: $\{5, 10, 50, 100, 400, 1500\}$ (Figure ~\ref{fig:ablation_MLP_size}). Overall, we find that increase in dimensionality leads to performance gains, with performance saturating when the dimension reaches over 50 units.

\section{Related work}
\textbf{Parameter-efficient tuning methods.} Recent approaches have explored parameter-efficient tuning (PEFT)  of language models, where only a subset of parameters is trained while the rest of the model is kept frozen. 
\citet{houlsby2019parameter} proposed \textit{adapters} -- small modules injected between each transformer layer. To improve over original adapter tuning, several works proposed to remove adapter modules from lower transformer layers \citep{ruckle2020adapterdrop}, or use a composition of adapter modules \citep{pfeiffer2020adapterfusion}. Other works focused on \textit{low-rank adaptations} (LoRA) \citep{hu2021lora}, and \textit{prefix tuning} \citep{li2021prefix}. Similarly to adapters, LoRA injects additional trainable weight matrices into each transformer layer, requiring changes to the intrinsic model structure and adding a high number of extra parameters. 

\textbf{Prompt tuning methods.} To overcome the drawbacks of traditional PEFT methods, \citet{lester2021power} introduced \textit{prompt tuning} as a highly parameter-efficient approach, where tuned soft prompts constitute $<0.1\%$ of the total parameters and can be easily appended to the input without modifying the model. Several methods were recently introduced to improve over prompt tuning. \citet{liu2021p} proposed adding soft prompts at every transformer layer. While their method improves performance, it requires much more trainable parameters (10x in some cases). Other works explored transfer learning-based methods to find better prompt initialization through pre-training \citep{vu2021spot, asai2022attentional}. These methods pre-train soft prompts on a collection of source tasks and subsequently use the learned prompt embeddings to initialize prompts for target tasks. 

\textbf{Reparameterization methods.} 
Although reparameterization has not been traditionally used with \textit{prompt tuning}, \citet{li2021prefix} explored \textit{reparameterization of embeddings} as a way to improve the performance of prefix tuning, and \citet{liu2021gpt} explored reparameterizing injectable embeddings together full model tuning. With these approaches, prefix embeddings are passed through a shallow neural network, such as MLP (in prefix tuning) or LSTM (in GPT2 tuning by \citet{liu2021gpt}), before being concatenated to the input embeddings (or representations) and passed into a subsequent layer of the language model. \citet{liu2021p} explores MLP-based reparameterization for \textit{P-tuning v2}. Despite improvements on some tasks, \citet{liu2021p} finds that the reparameterization effect is not consistent across datasets and can hinder the performance of certain tasks.

\section{Conclusion}
We propose \textsc{Residual Prompt Tuning}, a new method for learning soft prompts under a frozen language model using residual reparameterization of prompt embeddings. Our method enables efficient learning of soft prompts, without the need for extensive hyperparameter search, long training times, or pre-training on source tasks.  

Our experiments show that \textsc{Residual Prompt Tuning} significantly outperforms prompt tuning by \citet{lester2021power} and its two variations across three model architectures (T5-Large, T5-Base and BERT-Base) on SuperGLUE benchmark. Furthermore, our method is robust to the hyperparameter choice (learning rate and prompt initialization), speeds up convergence and is highly effective in few-shot settings.

\section*{Limitations}

Despite the simplicity and strong empirical results, \textsc{Residual Prompt Tuning} still has few limitations. First, its performance is still not on par with fine-tuning on (e.g. $7.8$ points difference with T5L model and 100-token prompt on SuperGLUE average score). Also, our method uses slightly more parameters than prompt tuning to train the reparameterization network. However, this is not a significant limitation given the full language model size. 
We have tried to cover several model architectures, but so far we have focused on encoder-decoder (T5) and encoder-only (BERT) models. In future work, we would like to investigate decoder-only methods (e.g. GPT).
Another limitation is that our method (similarly to other prompt tuning-based methods) strives to reduce the number of trainable parameters, but uses a longer sequence than the original input text (due to the injected prompt). 

\section*{Ethics Statement}
The main objective of \textsc{Residual Prompt Tuning} is to improve parameter-efficient tuning of large language models, which makes state-of-the-art models more accessible to groups with limited computational and data-labeling resources. We do not believe there is any potential risk in the published code or models in this work, as all of our experiments are based on public data that is widely used in the research community.

\section*{Acknowledgement}
We would like to thank Victoria Lin and extended FAIR team for their helpful feedback and comments. We also thank Akari Asai and Xiang Lisa Li for our discussions on the prompt tuning methodologies. 


\bibliography{anthology,custom}
\bibliographystyle{acl_natbib}

\appendix

\clearpage
{\large\textbf{Appendix}}
\section{Implementation and Training}
\subsection{Implementation details}
\label{appendix:implementation}
We use PyTorch \citep{paszke2019pytorch} and HuggingFace Transformers library \citep{wolf2019huggingface} for our implementation. To download data for SuperGLUE tasks, we use HuggingFace datasets (\url{https://github.com/huggingface/datasets})  \citep{wang2019superglue}. 

In our prompt tuning and reparameterization experiments, we follow setup from the previous works on prompt tuning \citep{lester2021power, vu2021spot}, and use the available validation set for each task to report the highest performance. 


\subsection{Datasets}
\label{appendix:datasets}
Table~\ref{tab:datasets_explained} shows details of the eight datasets from SuperGLUE benchmark \citep{wang2019superglue} that we used for our experiments, along with their training sizes and evaluation metrics. Following \citet{raffel2020exploring} and \citet{lester2021power}, for tasks that have two evaluation metrics we use the average of both scores as the final performance metric.

\begin{table}[h]
\scalebox{0.75}{
\addtolength{\tabcolsep}{-3pt} 
\begin{tabular}{l|llll}
\toprule
 \textbf{Dataset name} & \textbf{Train} & \textbf{Task} &  \textbf{Domain} & \textbf{Metric} \\
 \midrule
 1. BoolQ & 9,427 & QA & Wikipedia & acc. \\
 2. CB & 250 & NLI & various & F1 \& acc. \\
 3. COPA & 400 & QA & blogs, encyclop. & acc. \\
 4. MultiRC & 5,100 & QA & various & F1 \& EM \\
 5. ReCoRD & 101K & QA & various & F1 \& EM \\
 6. RTE & 2,500 & NLI & news, Wiki & acc. \\
 7. WiC & 6,000 & WSD & lexical databases & acc. \\
 8. WSC & 554/259\textsuperscript{*} & coref. & fiction books & acc.
 \\
    
 \bottomrule
\end{tabular}
}
\captionof{table}{The details of 8 SuperGLUE tasks used in our experiments. NLI denotes natural language inference, QA denotes questions and answers task, WSD denotes word sense disambiguation, EM denotes exact match scoring, acc. denotes accuracy. \textsuperscript{*}For T5 model variants we follow \citet{raffel2020exploring} setup, casting WSC as a text generation task and limit our training set to the positive examples only, where the supplied referent is correct (resulting in a total of 259 examples). For BERT we use the full WSC training set (554 examples) and train the model for binary classification, following \citet{wang2019superglue}.}
\label{tab:datasets_explained}
\end{table}

\subsection{Tokenization and Preprocessing}
\label{appendix:preprocessing}
Following common practice \citep{lester2021power, vu2021spot, asai2022attentional}, for all our experiments, we set the maximum input length (including the prepended prompt) to 512 tokens. We use padding to maximum length and mask out the padded tokens. In case of input exceeding 512 tokens, we truncate the input. We do not perform any specific text preprocessing (e.g. removing punctuation) but instead directly tokenize the raw text from SuperGLUE datasets using the corresponding model tokenizer from HuggingFace \citep{wolf2019huggingface}. 

For \textbf{BERT} experiments, we follow \citet{devlin2018bert} formatting -- the input sequence begins with $\mathtt{[CLS]}$ token, and ends with $\mathtt{[EOS]}$ token. For tasks with sentence pairs (e.g. RTE), we only insert our soft prompt before the first sentence, and concatenate both sentences with $\mathtt{[SEP]}$ token in between.

For \textbf{T5} experiments, we follow \citet{raffel2020exploring} formatting. We feed input examples along with their descriptors (e.g. "sentence1" and "sentence2"), and cast all classification tasks into text-to-text format (e.g. 0 and 1 classes in BoolQ task are cast into "True" and "False") replicating guidelines from \citet{raffel2020exploring}.

\subsection{Prompt initialization}
\label{appendix:prompt_init}
In all our experiments, unless otherwise specified, we initialize prompt virtual tokens using randomly sampled vocabulary embeddings \citep{lester2021power}. We sample uniformly across the whole vocabulary, without limiting to top-$k$ most common tokens. 
For our studies on performance robustness to the prompt initialization (Section~\ref{section:init}), we also explore random initialization, where embedding values are sampled uniformly from $[-0.5, 0.5]$ following \citet{lester2021power}.

\subsection{Training details}
\label{appendix:training_details}

\subsubsection{Infrastucture}
All of our experiments were conducted with 12 GPUs, with 32 GB memory each. On
each task, training took between 20 minutes and 26 hours.

\subsubsection{Hyperparameters}
Following \citet{lester2021power, vu2021spot}, we tune each method with a flat learning rate (LR) determined by hyperparameter search. Hyperparameter search was done via manual tuning and settings were selected based on the best SuperGLUE score (we use a subset of 5 tasks as in \citet{asai2022attentional}). 

For T5 models, we search LRs from $\{0.01, 0.1, 0.3, 0.7, 1.0\}$; based on the search use the following LRs: $0.7$ for \textsc{Residual Prompt Tuning}, MLP and LSTM-reparameterized prompt tunings, $0.3$ for the original prompt tuning (this also agrees with \citet{vu2021spot}). 

For BERT model, we search LRs from $\{10^{-6}, 5 \times 10^{-6}, 10^{-5}, 2 \times 10^{-5}, 5 \times 10^{-5}, 10^{-4}\}$; we find LR of $2 \times 10^{-5}$ to achieve the best performance with \textsc{Residual Prompt Tuning} and all prompt tuning variations, and use LR of $10^{-6}$ for fine-tuning according to \citet{wang2019superglue}.  

In all our experiments, we use batch size of 8 and AdamW optimizer \citep{loshchilov2018fixing} with the following hyperparameters: $\beta_1$ of 0.9, $\beta_2$ of 0.999, weight decay of 0.01, $\epsilon$ of $10^{-8}$ and bias correction turned on.

\subsubsection{MLP and LSTM design}
For \textsc{Residual Prompt Tuning} and prompt tuning w/ MLP we use two-layer MLP as shown in Figure~\ref{fig:mlp}. The only design difference between \textsc{Residual Prompt Tuning} and prompt tuning w/ MLP is the residual connection. We set the hidden layer dimension of MLP to 250 in parameter-efficient experiments (Section~\ref{section:subsection_peft}), and to 400 in all other experiments. We use ReLU non-linearity and apply LayerNorm normalization.

For prompt tuning w/ LSTM we use one-layer bidirectional LSTM with embedding dimension of 300, and dropout of 0.05, following \citet{liu2021gpt}.

\subsubsection{Training and evaluation}
We train all prompt tuning-based methods for 15 epochs in case of 10-token prompts and for 20 epochs in case of 100-token prompts. We run fine-tuning experiments for 30 epochs. 

In Section~\ref{section:subsection_peft}, where we compare parameter-efficient methods, we replicate training setup from \citet{asai2022attentional}, and trained our method for 20 epochs (since explored datasets are small-sized and contained less than 10k
examples) 

Since SuperGLUE tasks that we used in our study do not have a test set, we used validation set performance as a final performance metric, following previously used prompt tuning protocols by \citet{lester2021power} and \citet{vu2021spot}. We checkpoint the models every epoch, and report the highest validation performance. Similarly to \citet{lester2021power}, for each task we used its recommended metric by \citet{wang2019superglue} (see Table~\ref{tab:datasets_explained}); for tasks with two corresponding metrics we report the average of both scores.

\subsection{Parameter-efficiency of \textsc{Residual Prompt Tuning}}
\label{appendix:efficiecy}
The total number of trainable parameters in \textsc{Residual Prompt Tuning} consists of \textit{1)} trainable prompt embeddings, and \textit{2)} reparameterization network, which tunes down-projection $\mathbf{W}_{\text{down}} \in \mathbb{R}^{d \times m}$ and up-projection $\mathbf{W}_{\text{up}} \in \mathbb{R}^{m \times d}$ layers, as well as LayerNorm parameters (as shown in Figure~\ref{fig:mlp}). We assume that $d$ is the dimensionality of model embeddings, $m$ is MLP bottleneck size and $N$ is the number of prompt tokens. Hence, we have $d \times N$ soft prompt parameters, and 
$m \times d + d \times m + 2d = 2dm + 2d$ parameters in the reparameterization network. Thus, \textsc{Residual Prompt Tuning} has $2dm + 2d + dN$ trainable parameters. Importantly, the reparameterization network can be discarded after training, hence we only have $dN$ task-specific parameters.

\section{Performance on SuperGLUE}

\subsection{Performance with 100-token prompts}
\label{appendix:table100token}
Table~\ref{tab:100tokens} shows the performance of different approaches for prompt tuning (w/ and w/o reparameterization) with 100-token prompts, presenting per-task results for all SuperGLUE tasks across three model variants (T5-Large, T5-Base, BERT-Base). We see that our method, \textsc{Residual Prompt Tuning}, leads to consistent performance improvement over prompt
tuning and two reparameterization methods across different tasks.

\subsection{Convergence of different prompt tuning approaches}
\label{appendix:pt_res_mlp}
\begin{figure*}[h]
\centering
\includegraphics[scale=0.5]{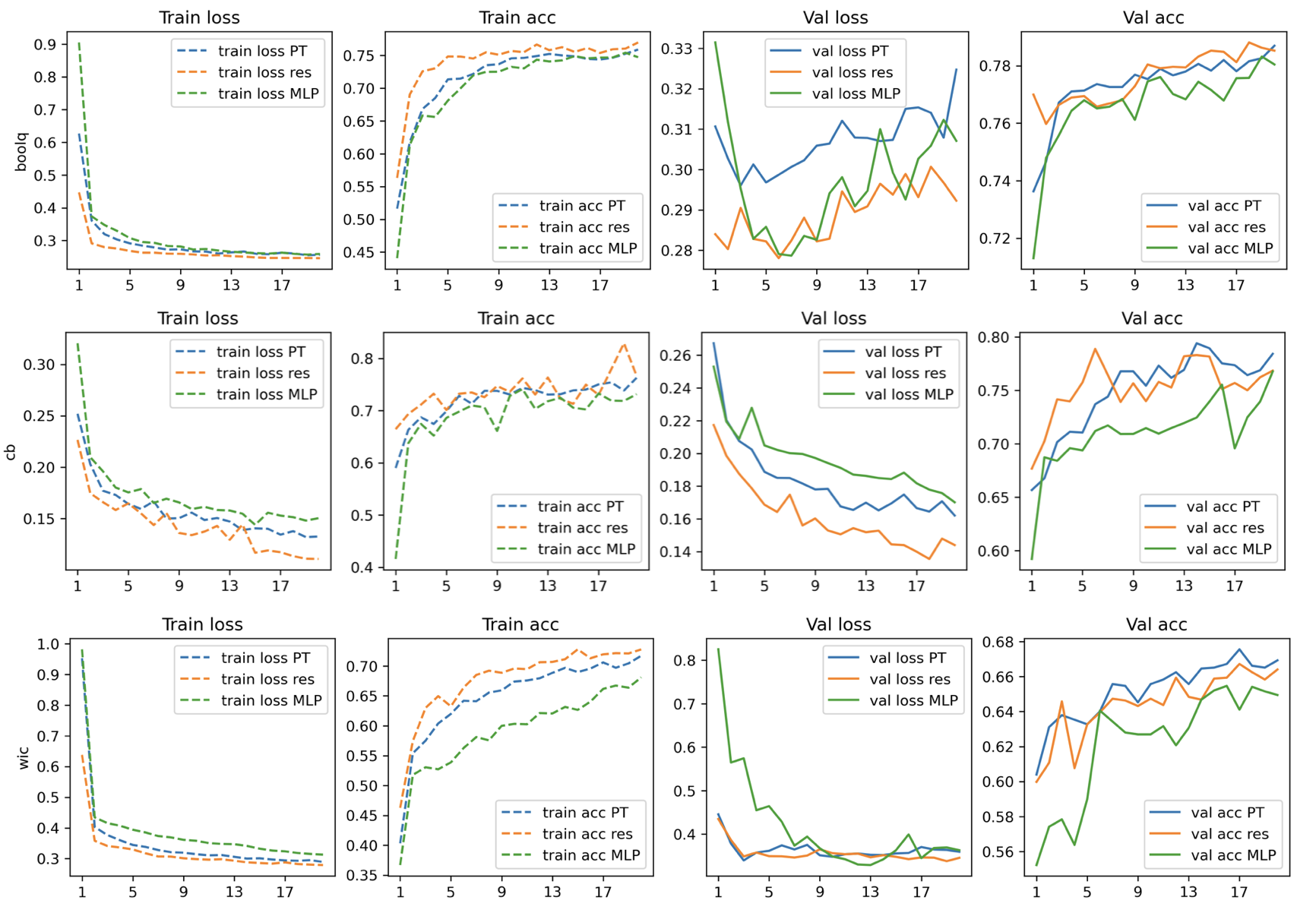}
\caption{Comparison of convergence of \textit{\textsc{Residual Prompt Tuning}} (in orange), \textit{prompt tuning} (in blue) and prompt tuning with MLP reparameterization (in green). Experiments are performed with T5-Base model and 10-token prompt. We show accuracy and loss on train and development sets over the course of training (20 epochs); each point on the plot is an average of 3 runs. Results are shown for BoolQ task (\textit{top plot}), CB task (\textit{middle plot}) and WiC task (\textit{bottom plot}). }
\label{fig:pt_res_mlp}
\end{figure*}

Here, we study the convergence of \textsc{Residual Prompt Tuning}, prompt tuning, and prompt tuning with MLP reparameterization. We show the evolution of accuracy and loss over the course of training on several SuperGLUE tasks in Figure~\ref{fig:pt_res_mlp}. We observe that \textsc{Residual Prompt Tuning} substantially speeds up convergence over the original prompt tuning by \citet{lester2021power}. Notably, the residual connection in the reparameterization network plays a key role in boosting performance -- MLP-based reparameterization without skip connection is actually slower to converge than the standard prompt tuning (Figure~\ref{fig:pt_res_mlp}). We hypothesize that this is explained by skip connection making it easier to optimize prompt embeddings. Specifically, skip connection allows to bypass learning the identity function, and learns projections "on top" of the original embeddings instead of learning them from scratch (similar observations by  \citep{he2016deep}). Thus, residual prompt repameterization allows to flexibly combine the original prompt embeddings with embeddings projections, resulting in faster convergence and improved performance. 



\section{Extended ablation studies}

\subsection{Effect of shared reparameterization network}
\label{appendix:ablations_shared}
\begin{figure*}[t]
\centering
\includegraphics[scale=0.42]{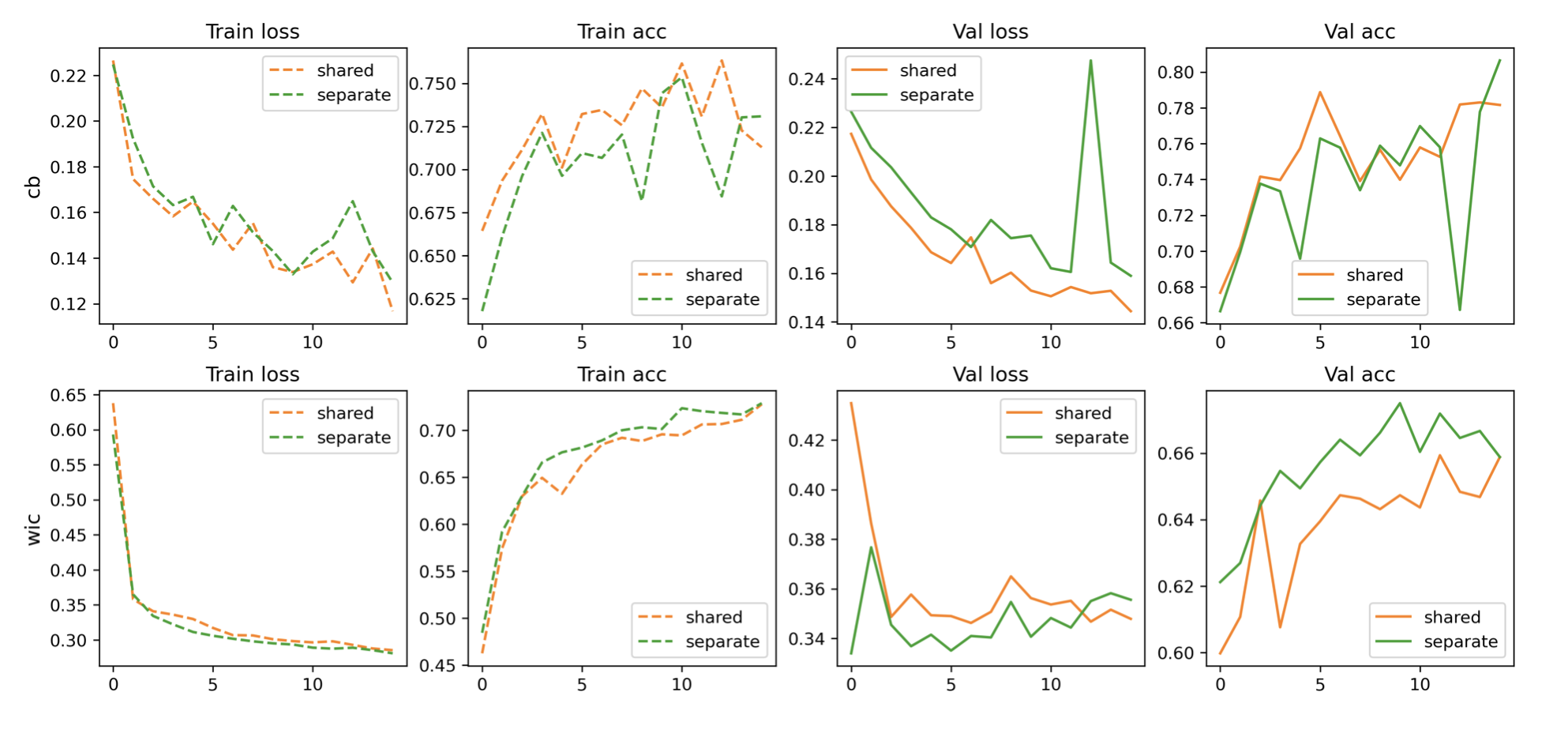}
\caption{Effect of \textit{shared} (in orange) versus \textit{separate} (in green) reparameterization network on the performance of \textsc{Residual Prompt Tuning}. Experiments are performed with T5-Base model and 10-token prompt. We show accuracy and loss on train and development sets over the course of training (20 epochs); each point on the plot is an average of 3 runs. Results are shown for CB task (\textit{top plot}) and WiC task (\textit{bottom plot}). }
\label{fig:shared_separate_mlp}
\end{figure*}

Figure~\ref{fig:shared_separate_mlp} shows performance of \textsc{Residual Prompt Tuning} with shared and separate MLP for reparameterization. Interestingly, shared MLP offers better performance on small datasets (e.g. CB) due to knowledge sharing between prompt tokens. At the same time, separate MLPs offer more flexibility and perform better on larger-scale datasets (e.g. WiC). Overall, their performance is similar and shared MLP is a significantly more parameter-efficient variant. Hence, we choose to use shared MLP in our work.

\label{appendix:ablations_nonlin}

\begin{table*}[h]
\newcolumntype{g}{>{\columncolor{black!15!white}}c}
\setlength\extrarowheight{-2.1pt}
  \centering
  \scalebox{0.79}{
  \begin{tabular}{lccccccccg}
  \toprule
   Task $\rightarrow$  & BoolQ & CB & COPA & MultiRC & ReCoRD & RTE & WiC & WSC & Avg. \\
   Method $\downarrow$  & Acc.  & F1/Acc.  & Acc.  & F1/EM  & F1/EM & Acc. & Acc.  & Acc. & - \\
\toprule \multicolumn{10}{c}{\textbf{T5-Large}} \\

Prompt Tuning\textsuperscript{\ddag} & - & - & - & - & - & - & - & - & \textbf{74.5} \\
PT w/ MLP & 83.7 &  87.1 &  52.7 & 65.3 & 77.4 & 85.7 &  68.5 &  21.9 &  67.8 \\
Residual PT & 84.2 &  93.3 &  54.3 & 83.9 & 65.9 & 87.7 &  71.1 &  55.3 & \textbf{74.5} \\
\midrule
Fine-tuning\textsuperscript{\dag} &  85.4 & 93.2 & 83.4 & 67 & 86.3 & 87.8 & 69.3 & 86.3 & 82.3 \\
\toprule \multicolumn{10}{c}{\textbf{T5-Base}} \\

Prompt Tuning\textsuperscript{\ddag} & - & - & - & - & - & - & - & - & 63.1 \\
PT w/ MLP &   72.7 &  78.7 &  56.3 &     58.1 &    63.0 & 61.4 &  66.1 &  43.0 &  62.4 \\
Residual PT  &   79.0 &  86.0 &  60.0 &     79.6 &    56.7 & 81.5 &  68.4 &  52.6 &  \textbf{70.5}  \\

\midrule
Fine-tuning\textsuperscript{\dag} & 81.4 & 86.2 & 94.0 & 71.2 & 61.4 & 74.6 & 68.3 & 80.8 & 76.2 \\
\toprule \multicolumn{10}{c}{\textbf{BERT-Base}} \\

Prompt Tuning &   62.2 &  62.5 &  54.6 &     57.4 & 64.8 &  52.5 &  55.4 &  64.1 &  59.2 \\
PT w/ MLP &   62.3 &  63.7 &  64.0 &     58.3 &    65.2 &  51.5 &  57.1 &  64.4 &  60.8 \\
Residual PT &   62.3 &  72.6 &  64.2 &  57.8 &    65.2 &  52.7 &  54.2 &  63.8 &  \textbf{61.6}  \\
\midrule
Fine-tuning & 73.2 &  89.9 &  65.7 & 66.9 &    62.8 &  65.1 &  67.8 &  63.8 &  69.4 \\
  \bottomrule
  \end{tabular}
  }
  \caption{Results on SuperGLUE development set with \textbf{100-token prompt}. All scores are averaged over 3 runs. \textsuperscript{\ddag}denotes results reported by \citet{vu2021spot, lester2021power} (only average SuperGLUE performance is reported). \textsuperscript{\dag}denotes results reported by \citet{raffel2020exploring}. For tasks with two corresponding scores the average of both scores is reported.}
  \label{tab:100tokens}
\end{table*}

\end{document}